\documentclass[sigconf]{acmart}
\copyrightyear{2020}
\acmYear{2020}
\acmConference[WWW '20]{Proceedings of The Web Conference 2020}{April 20--24, 2020}{Taipei, Taiwan}
\acmBooktitle{Proceedings of The Web Conference 2020 (WWW '20), April 20--24, 2020, Taipei, Taiwan}
\acmPrice{}
\acmDOI{10.1145/3366423.3380191}
\acmISBN{978-1-4503-7023-3/20/04}

\usepackage{soul}
\usepackage{graphicx}
\usepackage{amssymb}
\usepackage{subfigure}
\usepackage{color}
\usepackage{multirow}

\newcommand{\ie}{\emph{i.e.,}}

\newcommand{\ignore}[1]{}

\newcommand{\dubbelop}{$^{\blacktriangle}$}

\newcommand{\dubbelneer}{$^{\blacktriangledown}$}

\begin{document}
\title{Learning to Respond with Stickers: \\ A Framework of Unifying Multi-Modality in Multi-Turn Dialog}

\author{Shen Gao}
\authornote{Equal contribution. Ordering is decided by a coin flip. Work performed during an internship at IIAI.}
\authornote{WICT is the abbreviation of Wangxuan Institute of Computer Technology.}
\affiliation{%
  \institution{WICT, Peking University}
}
\email{shengao@pku.edu.cn}

\author{Xiuying Chen}
\authornotemark[1]
\affiliation{%
  \institution{WICT, Peking University}
}
\email{xy-chen@pku.edu.cn}

\author{Chang Liu}
\affiliation{%
  \institution{WICT, Peking University}
}
\email{liuchang97@pku.edu.cn}

\author{Li Liu}
\affiliation{%
  \institution{Inception Institute of Artificial Intelligence}
}
\email{li-liu1985@inceptioniai.org}

\author{Dongyan Zhao}
\affiliation{%
	\institution{WICT, Peking University}
}
\email{zhaody@pku.edu.cn}

\author{Rui Yan}
\authornote{Corresponding Author: Rui Yan (ruiyan@pku.edu.cn)}
\affiliation{
  \institution{\textsuperscript{1} WICT, Peking University\\
  \textsuperscript{2} Beijing Academy of Artificial Intelligence}
}
\email{ruiyan@pku.edu.cn}

\renewcommand{\shortauthors}{Shen and Xiuying, et al.}

\begin{abstract}
Stickers with vivid and engaging expressions are becoming increasingly popular in online messaging apps, and some works are dedicated to automatically select sticker response by matching text labels of stickers with previous utterances.
However, due to their large quantities, it is impractical to require text labels for the all stickers.
Hence, in this paper, we propose to recommend an appropriate sticker to user based on multi-turn dialog context history without any external labels.
Two main challenges are confronted in this task.
One is to learn semantic meaning of stickers without corresponding text labels.
Another challenge is to jointly model the candidate sticker with the multi-turn dialog context.
To tackle these challenges, we propose a \emph{sticker response selector} (SRS) model.
Specifically, SRS first employs a convolutional based sticker image encoder and a self-attention based multi-turn dialog encoder to obtain the representation of stickers and utterances.
Next, deep interaction network is proposed to conduct deep matching between the sticker with each utterance in the dialog history.
SRS then learns the short-term and long-term dependency between all interaction results by a fusion network to output the the final matching score.
To evaluate our proposed method, we collect a large-scale real-world dialog dataset with stickers from one of the most popular online chatting platform.
Extensive experiments conducted on this dataset show that our model achieves the state-of-the-art performance for all commonly-used metrics.
Experiments also verify the effectiveness of each component of SRS.
To facilitate further research in sticker selection field, we release this dataset of 340K multi-turn dialog and sticker pairs\footnote{\url{https://github.com/gsh199449/stickerchat}}.
\end{abstract}


%

\maketitle

\section{Introduction}
\label{sec:intro}
\begin{figure}
    \centering
    \includegraphics[scale=0.13]{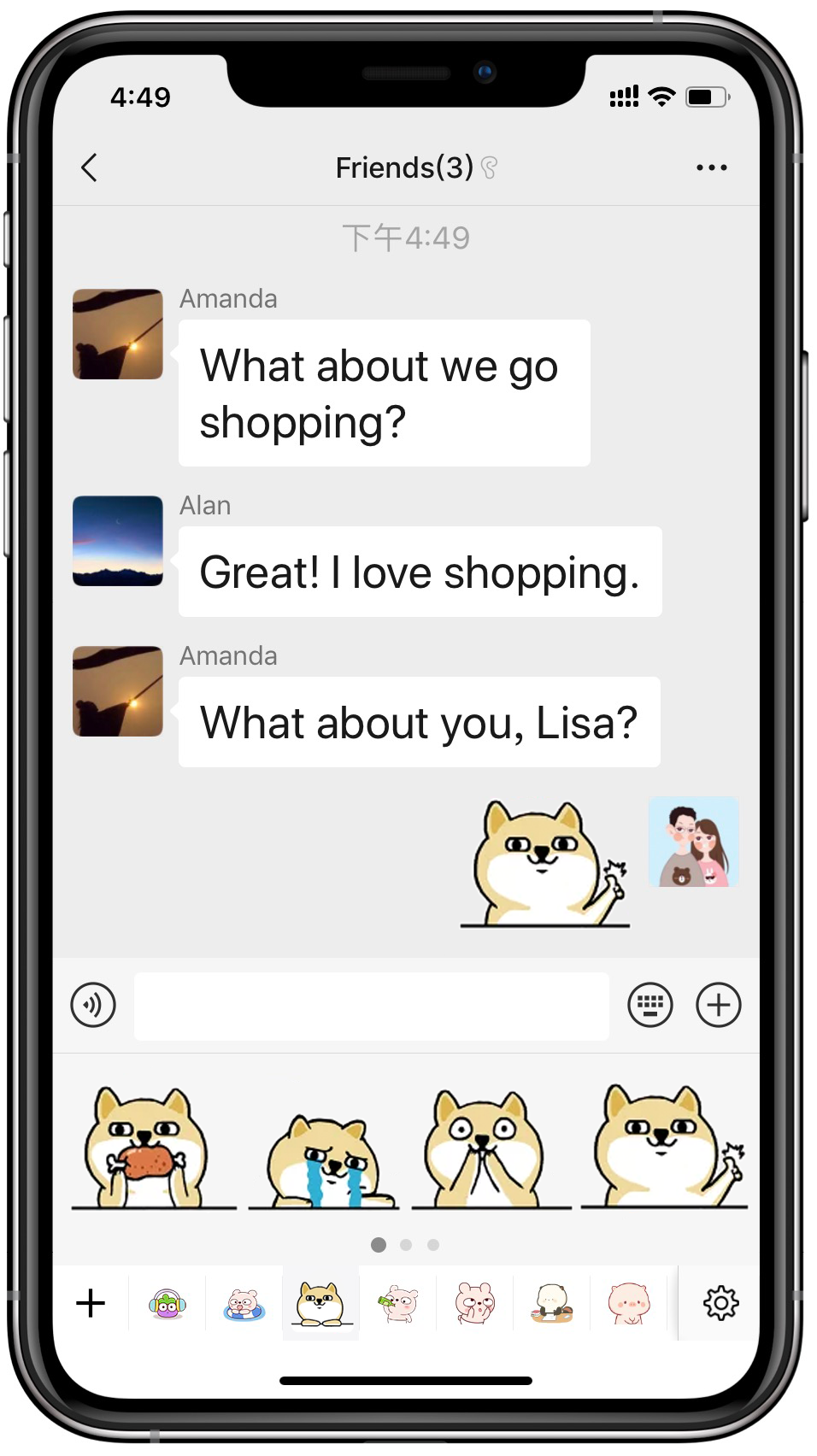}
    \caption{
         An example of stickers in a multi-turn dialog. Sticker response selector automatically selects the proper sticker based on multi-turn dialog history. 
    }
    \label{fig:example}
\end{figure}

Images are another important approach for expressing feelings and emotions in addition to using text in communication.
In mobile messaging apps, these images can generally be classified into emojis and stickers.
Emojis are usually used to help reinforce simple emotions in a text message due to their small size, and their variety is limited.
Stickers, on the other hand, can be regarded as an alternative for text messages, which usually include cartoon characters and are of high definition.
They can express much more complex and vivid emotion than emojis.
Most messaging apps, such as WeChat, Telegram, WhatsApp, and Slack provide convenient ways for users to download stickers for free, or even share self-designed ones.
We show a chat window including stickers in Figure~\ref{fig:example}. 

Stickers are becoming more and more popular in online chat.
First, sending a sticker with a single click is much more convenient than typing text on the 26-letter keyboard of a small mobile phone screen.
Second, there are many implicit or strong emotions that cannot be accurately explained by words but can be captured by stickers with vivid facial expressions and body language.
However, the large scale use of stickers means that it is not always straightforward to think of the sticker that best expresses one's feeling according to the current chatting context.
Users need to recall all the stickers they have collected and selected the appropriate one, which is both difficult and time-consuming.

Consequently, much research has focused on recommending appropriate emojis to users according to the chatting context.
Existing works, such as~\cite{xie2016neural}, are mostly based on emoji recommendation, where they predict the probable emoji given the contextual information from multi-turn dialog systems.
In contrast, other works~\cite{barbieri2017emojis,barbieri2018multimodal} recommend emojis based on the text and images posted by a user.
However, the use of emojis is restricted due to their limited variety and small size, while stickers are more expressive and of a great variety.
As for sticker recommendation, existing works such as~\cite{laddha2019understanding} and apps like Hike or QQ directly match the text typed by the user to the short text tag assigned to each sticker.
However, since there are countless ways of expressing the same emotion, it is impossible to capture all variants of an utterance as tags.

In this paper, we address the task of sticker response selection in multi-turn dialog, where an appropriate sticker is recommended based on the dialog history.
There are two main challenges in this task:
(1) 
To the best of our knowledge, no existing image recognition methods can model the sticker image,
how to capture the semantic meaning of sticker is challenging.
(2) Understanding multi-turn dialog history information is crucial for sticker recommendation, and jointly modeling the candidate sticker with multi-turn dialog is challenging.
Herein, we propose a novel sticker recommendation model, namely \emph{sticker response selector} (SRS), for sticker response selection in multi-turn dialog.
Specifically, SRS first learns representations of dialog context history using a self-attention mechanism and learns the sticker representation by a convolutional network. 
Next, SRS conducts deep matching between the sticker and each utterance and produces the interaction results for every utterance.
Finally, SRS employs a fusion network which consists of a sub-network fusion RNN and fusion transformer to learn the short and long term dependency of the utterance interaction results.
The final matching score is calculated by an interaction function.
To evaluate the performance of our model, we propose a large number of multi-turn dialog dataset associated with stickers from one of the popular messaging apps. 
Extensive experiments conducted on this dataset show that SRS significantly outperforms the state-of-the-art baseline methods in commonly-used metrics.

\noindent 
\noindent Our contributions can be summarized as follows:



$\bullet$ We employ a deep interaction network to conduct matching between candidate sticker and each utterance in dialog context.

$\bullet$ We propose a fusion network that can capture the short and long dependency of the interaction results of each utterance simultaneously.

$\bullet$ Experiments conducted on a large-scale real-world dataset\footnote{https://github.com/gsh199449/stickerchat} show that our model outperforms all baselines, including state-of-the-art models.


\section{Related Work}
\label{sec:related}

We outline related work on sticker recommendation, visual question answering, visual dialog, and multi-turn response selection.

\textbf{Sticker recommendation.}
Most of the previous works emphasize the use of emojis instead of stickers.
For example, \cite{barbieri2017emojis, barbieri2018multimodal} use a multimodal approach to recommend emojis based on the text and images in an Instagram post.
However, emojis are typically used in conjunction with text, while stickers are independent information carriers.
What is more, emojis are limited in variety,  while there exists an abundance of different stickers.
The most similar work to ours is \cite{laddha2019understanding}, where they generate recommended stickers by first predicting the next message the user is likely to send in the chat, and then substituting it with an appropriate sticker.
However, more often than not the implication of the stickers cannot be fully conveyed by text and, in this paper, we focus on directly generating sticker recommendations from dialog history.

\textbf{Visual question answering.}
Sticker recommendation involves the representation of and interaction between images and text, which is related to the Visual Question Answering (VQA) task~\cite{Goyal2018Think,Gao2019Multi,Chao2018Cross,Wang2017Explicit,Noh2019Transfer}.
Specifically, VQA takes an image and a corresponding natural language question as input and outputs the answer.
It is a classification problem in which candidate answers are restricted to the most common answers appearing in the dataset and requires deep analysis and understanding of images and questions such as image recognition and object localization~\cite{malinowski2015ask,xiong2016dynamic,wu2016ask,goyal2017making}.
Current models can be classified into three main categories: early fusion models, later fusion models, and external knowledge-based models.
One state-of-the-art VQA model is \cite{li2019beyond}, which proposes an architecture, positional self-attention with co-attention, that does not require a recurrent neural network (RNN) for video question answering.
\cite{guo2019image} proposes an image-question-answer synergistic network, where candidate answers are coarsely scored according to their relevance to the image and question pair in the first stage. 
Then, answers with a high probability of being correct are re-ranked by synergizing with images and questions.

The difference between sticker selection and VQA task is that sticker selection task focus more on multi-turn multimodal interaction between stickers and utterances.

\textbf{Visual dialog.}
Visual dialog extends the single turn dialog task~\cite{Tao2018Get,Gao2019Product} in VQA to a multi-turn one, where later questions may be related to former question-answer pairs.
To solve this task, \cite{lu2017best} transfers knowledge from a pre-trained discriminative network to a generative
network with an RNN encoder, using a perceptual loss.
\cite{wu2018you} combines reinforcement learning and generative adversarial networks (GANs) to generate more human-like responses to questions, where the GAN helps overcome the relative paucity of training data, and the
tendency of the typical maximum-likelihood-estimation-based approach to generate overly terse answers.
\cite{jain2018two} demonstrates a simple symmetric discriminative baseline that can be applied to both predicting an answer as well as predicting a question in the visual dialog.

Unlike VQA and visual dialog tasks, in a sticker recommendation system, the candidates are stickers rather than text.

\textbf{Multi-turn response selection.}
Multi-turn response selection~\cite{Tao2019One,Feng2019Learning,Yan2018Coupled,Yan2017Joint,Yan2016LearningTR} takes a message and utterances in its previous turns as input and selects a response that is natural and relevant to the whole context.
In our task, we also need to take previous multi-turn dialog into consideration.
Previous works include \cite{zhou2016multi}, which uses an RNN to represent context and response, and measure their relevance.
More recently, \cite{Wu2017SequentialMN} matches a response with each utterance in the context on multiple levels of granularity, and the vectors are then combined through an RNN.
The final matching score is calculated by the hidden states of the RNN.
\cite{zhou2018multi} extends this work by considering the matching with dependency information.
More recently, \cite{tao2019multi} proposes a multi-representation fusion network where the representations can be fused into matching at an early stage, an intermediate stage, or at the last stage.

Traditional multi-turn response selection deals with pure natural language processing, while in our task, we also need to obtain a deep understanding of images.
\section{Dataset}
\label{sec:dataset}

In this section, we introduce our multi-turn dialog dataset with sticker as response in detail.

\subsection{Data Collection}

We collect the large-scale multi-turn dialog dataset with stickers from one of the most popular messaging apps.
In this app, a large mount of sticker sets are published, and everyone can use the sticker when chatting with a friend or in a chat group.
Specifically, we select 20 public chat groups consisting of active members, which are all open groups that everyone can join it without any authorities.
The chat history of these groups is collected along with the complete sticker sets.
These sticker sets include stickers with similar style.
All stickers are resized to a uniform size of $128 \times 128$ pixels.
We use 20 utterances before the sticker response as the dialog context, and then we filter out irrelevant utterance sentences, such as URL links and attached files.
Due to privacy concern, we also filter out user information and anonymize user IDs.
To construct negative samples, 9 stickers other than the ground truth sticker are randomly sampled from the sticker set.
After pre-processing, there are 320,168 context-sticker pairs in the training dataset, 10,000 pairs in the validation, and 10,000 pairs in test datasets respectively.
We make sure that there is no overlap between these three datasets.
Two examples are shown in Figure~\ref{fig:dataset-case}.
We publish this dataset to communities to facilitate further research on dialog response selection task.

\subsection{Statistics and Analysis}

\begin{table}[t]
\centering
\caption{Statistics of Response Selection Dataset.}
\label{tab:stat-dataset}
\begin{tabular}{llll}
\toprule
 & Train & Valid & Test \\
\midrule
\# context-stickers pairs & 320,168 & 10,000 & 10,000 \\
Avg. words of context utterance & 7.54 & 7.50 & 7.42 \\
Avg. users participate & 5.81 & 5.81 & 5.79 \\
\bottomrule
\end{tabular}
\end{table}
In total, there are 3,516 sets of sticker which contain 174,695 stickers.
The average number of stickers in a sticker set is 49.64.
Each context includes 15.5 utterances on average.
The average number of users who participate in the dialog context over each dataset is shown in the third row of Table~\ref{tab:stat-dataset}.

\subsection{Sticker Similarity}

\begin{figure}[t]
    \centering
    \includegraphics[scale=0.40]{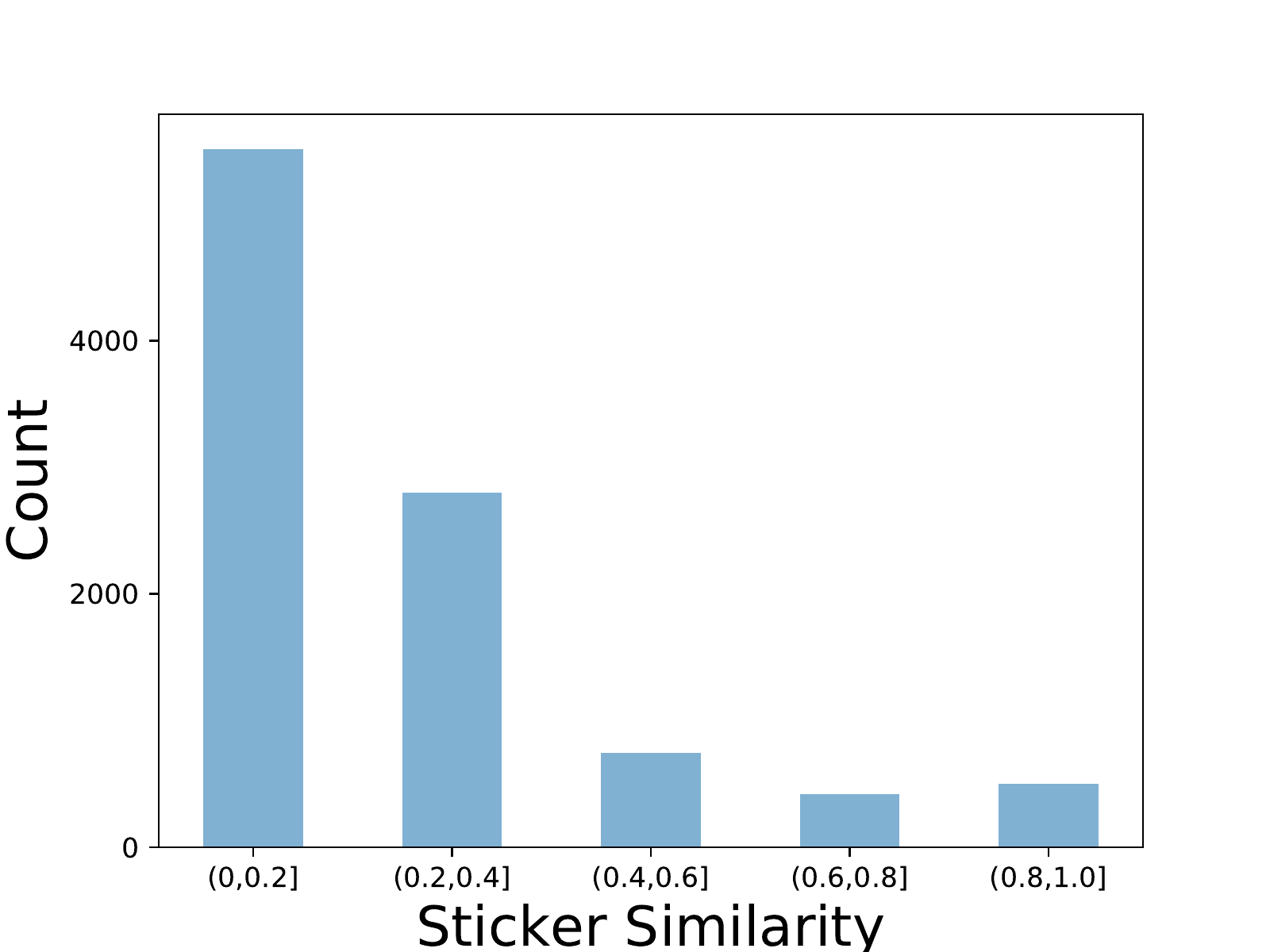}
    \caption{
        Similarity distribution among all stickers in test dataset.
    }
    \label{fig:similarity-distribution}
\end{figure}

\begin{figure*}[t]
    \centering
    \includegraphics[scale=0.58]{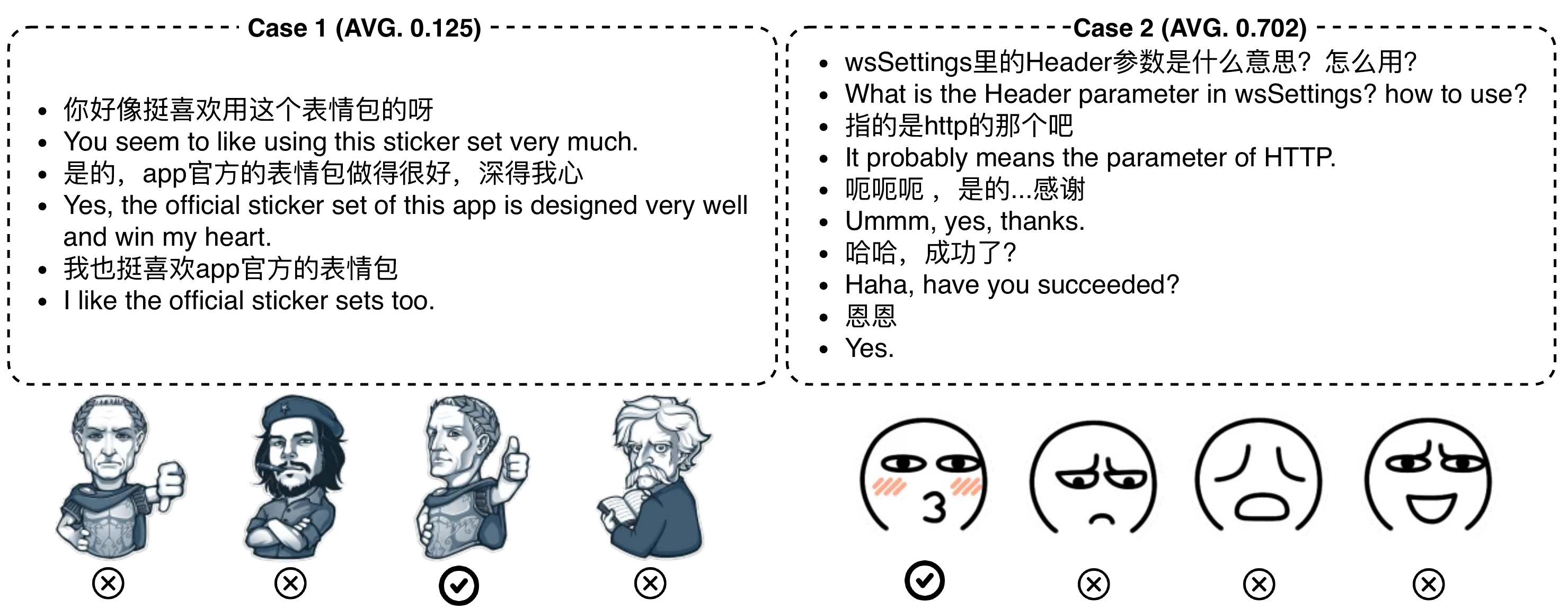}
    \caption{
        Example cases in the dataset with different similarity scores.
    }
    \label{fig:dataset-case}
\end{figure*}

Stickers in the same set always share a same style or contain the same cartoon characters.
Intuitively, the more similar the candidate stickers are, the more difficult it is to choose the correct sticker from candidates.
In other words, the similarity between candidate stickers determines the difficulty of the sticker selection task.
To investigate the difficulty of this task, we calculate the average similarity of all the stickers in a specific sticker set by the Structural Similarity Index (SSIM) metric~\cite{wang2004image,avanaki2008exact}.
We first calculate the similarity between the ground truth sticker and each negative sample, then average the similarity scores.
The similarity distribution among test data is shown in Figure~\ref{fig:similarity-distribution}, where the average similarity is 0.258.
The examples in Figure~\ref{fig:dataset-case} are also used to illustrate the similarity of stickers more intuitively, where the left one has a relatively low similarity score, and the right one has a high similarity score.

\section{Problem formulation}
\label{sec:formulation}

Before presenting our approach for sticker response selection in multi-turn dialog, we first introduce our notations and key concepts. 


Similar to the multi-turn dialog response selection~\cite{Wu2017SequentialMN,zhou2018multi}, we assume that there is a multi-turn dialog context $s=\{u_{1},\dots,u_{T_u}\}$ and a candidate sticker set $C=\{c_{1},...c_{T_c}\}$, where $u_{i}$ represents the $i$-th utterance in the multi-turn dialog.
In the $i$-th utterance $u_i=\{x^i_1,\dots,x^{i}_{T_x^i}\}$, $x^i_j$ represents the $j$-th word in $u_i$, and $T_x^i$ represents the total number of words in $u_i$ utterance.
In dialog context $s$, $c_{i}$ represents a sticker image with a binary label $y_i$, indicating whether $c_i$ is an appropriate response for $s$.
$T_u$ is the utterance number in the dialog context and $T_c$ is the number of candidate stickers.
For each candidate set, there is only one ground truth sticker, and the remaining ones are negative samples.
Our goal is to learn a ranking model that can produce the correct ranking for each candidate sticker $c_i$; that is, can select the correct sticker among all the other candidates.
For the rest of the paper, we take the $i$-th candidate sticker $c_{i}$ as an example to illustrate the details of our model and omit the candidate index $i$ for brevity.

\section{SRS model}
\label{section4}

\begin{figure*}
    \centering
    \includegraphics[scale=0.8]{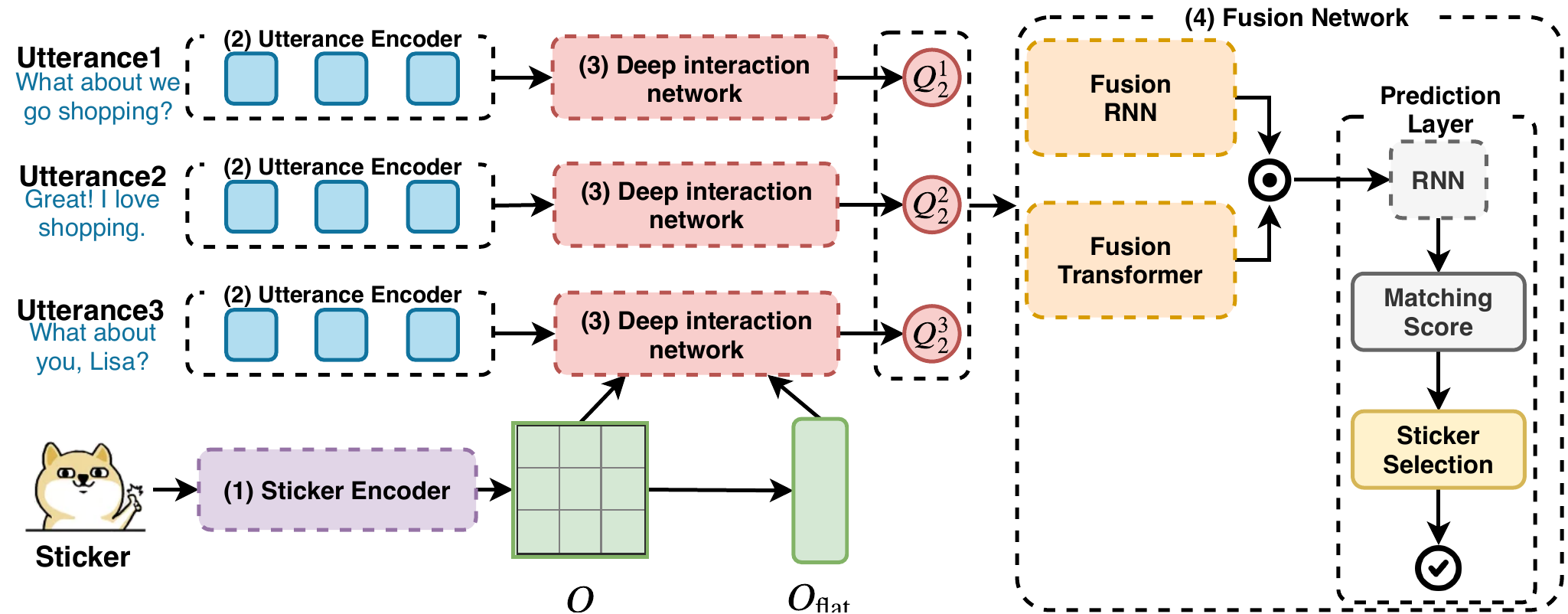}
    \caption{
     Overview of SRS. 
     We divide our model into four ingredients: 
     (1) \textit{Sticker encoder} learns sticker representation; 
     (2) \textit{Utterance encoder} learns representation of each utterance; 
     (3) \textit{Deep interaction network} conducts deep matching interaction between sticker representation and utterance representation in different levels of granularity.
     (4) \textit{Fusion network} combines the long-term and short-term dependency feature between interaction results produced by (3).
    }
    \label{fig:model}
\end{figure*}

\subsection{Overview}

In this section, we propose our \emph{sticker response selector}, abbreviated as SRS. 
An overview of SRS is shown in Figure~\ref{fig:model}, which can be split into four main parts:

$\bullet$ \textit{Sticker encoder} is a convolutional neural network (CNN) based image encoding module that learns a sticker representation. 

$\bullet$ \textit{Utterance encoder} is a self-attention mechanism based module encoding each utterance $u_{i}$ in the multi-turn dialog context $s$.

$\bullet$ \textit{Deep interaction network} module conducts deep matching between each sticker representation and each utterance, and outputs each interaction result.

$\bullet$ \textit{Fusion network} learns the short-term dependency by the fusion RNN and the long-term dependency by the fusion Transformer, and finally outputs the matching score combining these features using an interaction function.

\subsection{Sticker Encoder}
\label{subsec:sticker_encoder}
Much research has been conducted to alleviate gradient vanishing~\cite{he2016deep} and reduce computational costs~\cite{he2015delving} in image modeling tasks.
We utilize one of these models, \ie the Inception-v3~\cite{szegedy2016rethinking} model rather than plain CNN to encode sticker image:
\begin{align}
   O, O_{\text{flat}} &= \text{Inception-v3}(c) , \label{eq:inceptionv3}
\end{align}
where $c$ is the sticker image.
The sticker representation is $O \in \mathbb{R}^{p \times p \times d}$ which conserves the two-dimensional information of the sticker, and will be used when associating stickers and utterances in \S\ref{deep_int}.
We use the original image representation output of Inception-v3 $O_{\text{flat}} \in \mathbb{R}^{d}$ as another sticker representation.
However, existing pre-trained CNN networks including Inception-v3 are mostly built on real-world photos.
Thus, directly applying the pre-trained networks on stickers cannot speed up the training process.
In this dataset, sticker author give each sticker $c$ an emoji tag which denotes the general emotion of the sticker.
Hereby, we propose an auxiliary sticker classification task to help the model converge quickly, which uses $O_{\text{flat}}$ to predict which emoji is attached to the corresponding sticker.
More specifically, we feed $O_{\text{flat}}$ into a linear classification layer and then use the cross-entropy loss $\mathcal{L}_s$ as the loss function of this classification task.

\subsection{Utterance Encoder}

To model the semantic meaning of the dialog context, we learn the representation of each utterance $u_i$.
First, we use an embedding matrix $e$ to map a one-hot representation of each word in each utterance $u_i$ to a high-dimensional vector space.
We denote $e(x^i_j)$ as the embedding representation of word $x^i_j$.
From these embedding representations, we use the attentive module from Transformer~\cite{vaswani2017attention} to model the temporal interactions between the words in an utterance.
Attention mechanisms have become an integral part of compelling sequence modeling in various tasks~\cite{bahdanau2014neural,fan2018hierarchical,Gao2019Abstractive,li2019beyond}.
In our sticker selection task, we also need to let words fully interact with each other to model the dependencies of words without regard to their locations in the input sentence.
The attentive module in the Transformer has three inputs: the query $Q$, the key $K$ and the value $V$.
We use three fully-connected layers with different parameters to project the embedding of dialog context $e(x^i_j)$ into three spaces:
\begin{align}
    Q^i_j=FC(e(x^i_j)),    K^i_j=FC(e(x^i_j)) ,
    V^i_j=FC(e(x^i_j)).
\end{align}
The attentive module then takes each $Q^i_j$ to attend to $K^i_\cdot$, and uses these attention distribution results $\alpha^{i}_{j, \cdot} \in \mathbb{R}^{T_x^i}$ as weights to gain the weighted sum of $V^i_j$ as shown in Equation~\ref{equ:transformer-sum}.
Next, we add the original word representations on $\beta^i_{j}$ as the residential connection layer, shown in Equation~\ref{equ:drop-add}:
\begin{align}
    \alpha^i_{j,k} &= \frac{\exp\left( Q^i_j \cdot K^i_k \right)}{\sum_{n=1}^{T_x^i} \exp\left(Q^i_j \cdot K^i_n\right)}, \label{equ:attention}\\
    \beta^i_{j} &= \textstyle \sum_{k=1}^{T_x^i} \alpha^i_{j,k} \cdot V^i_{k}, \label{equ:transformer-sum}\\
    \hat{h}^i_j &= \text{Dropout} \left( e(x^i_j) + \beta^i_j \right), \label{equ:drop-add}
\end{align}
where $\alpha^i_{j,k}$ denotes the attention weight between $j$-th word to $k$-th word in $i$-th utterance.
To prevent vanishing or exploding of gradients, a layer normalization operation~\cite{lei2016layer} is also applied on the output of the feed-forward layer  as shown in Equation~\ref{equ:ffn}: 
\begin{align}
    h^i_j &= \text{norm}\left(\max(0, \hat{h}^i_j \cdot W_1 + b_1) \cdot W_2 + b_2 \right), \label{equ:ffn}
\end{align}
where $W_1, W_2, b_1, b_2$ are all trainable parameters.
$h^{i}_j$ denotes the hidden state of $j$-th word in the Transformer for the $i$-th utterance. 

\subsection{Deep Interaction Network}
\label{deep_int}

\begin{figure*}
    \centering
    \includegraphics[scale=0.75]{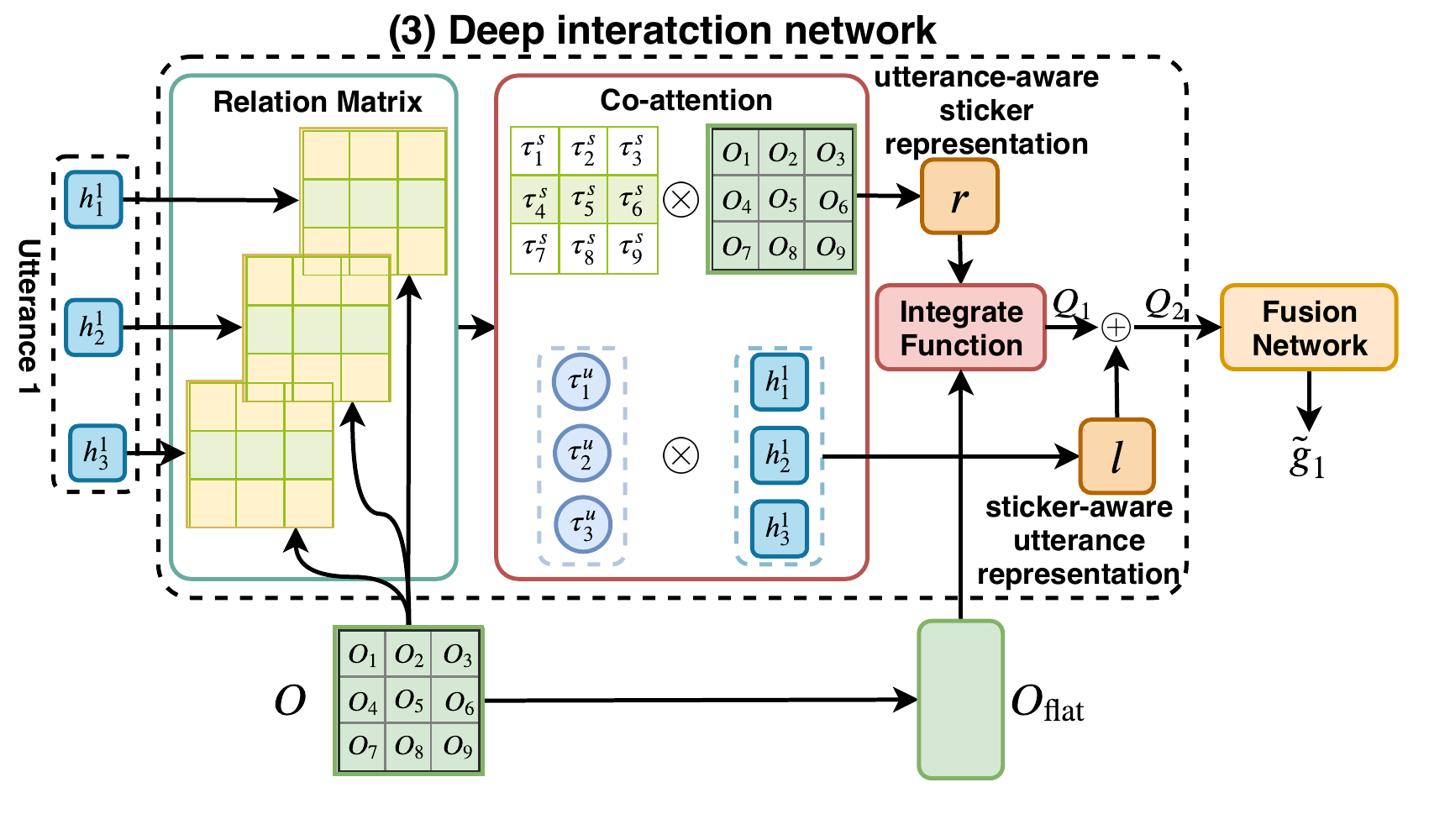}
    \caption{Framework of deep interaction network.}
    \label{fig:interaction}
\end{figure*}

Now that we have the representation of the sticker and each utterance, we can conduct a deep matching between these components.
On one hand, there are some emotional words in dialog context history that match the expression of the stickers such as ``happy'' or ``sad''.
On the other hand, specific part of the sticker can also match these corresponding words such as dancing limbs or streaming eyes.
Hence, we employ a bi-directional attention mechanism between a sticker and each utterance, that is, from utterance to sticker and from sticker to utterance, to analyze the cross-dependency between the two components.
The interaction is illustrated in Figure~\ref{fig:interaction}.

We take the $i$-th utterance as an example and omit the index $i$ for brevity.
The two directed attentions are derived from a shared relation matrix, $M \in  \mathbb{R}^{(p^2) \times  T_{u}}$, calculated by sticker representation $O \in \mathbb{R}^{p \times p \times d}$ and utterance representation $h \in \mathbb{R}^{T_{u} \times d}$.
The score $M_{kj} \in \mathbb{R}$ in the relation matrix $M$ indicates the relation between the $k$-th sticker representation unit $O_k$, $k \in [1,p^2]$ and the $j$-th word $h_j$, $j \in [1, T_{u}]$ and is computed as:
\begin{equation}
    M_{kj} = \sigma(O_k, h_j), \quad \sigma(x, y) = w^\intercal [x \oplus y \oplus (x \otimes y)] ,
\label{eq:alpha}
\end{equation}
where $\sigma$ is a trainable scalar function that encodes the relation between two input vectors. 
$\oplus$ denotes a concatenation operation and $\otimes$ is the element-wise multiplication.

Next, a max pooling operation is conducted on $M$, \ie let $\tau_j^u = \max(M_{:j}) \in \mathbb{R}$ represent the attention weight on the $j$-th utterance word by the sticker representation, corresponding to the ``utterance-wise attention''.
This attention learns to assign high weights to the important words that are closely related to sticker.
We then obtain the weighted sum of hidden states as ``\textbf{sticker-aware utterance representation}'' $l$:
\begin{equation}\label{equ:sa-utterance}
l = \textstyle \sum^{T_{u}}_j {\tau_j^u h_j} .
\end{equation}

Similarly, sticker-wise attention learns which part of a sticker is most relevant to the utterance.
Let $\tau_k^s=\text{max}(M_{k:}) \in \mathbb{R}$ represent the attention weight on the $k$-th unit of the sticker representation. We use this to obtain the weighted sum of $O_{k}$, \ie the ``\textbf{utterance-aware sticker representation}'' $r$:
\begin{equation}\label{equ:ua-sticker}
r = \textstyle \sum^{p^2}_k {\tau_k^s O_{k}} .
\end{equation}

After obtaining the two outputs from the co-attention module, we combine the sticker and utterance representations and finally get the ranking result.
We first integrate the utterance-aware sticker representation $r$ with the original sticker representation $O_{\text{flat}}$ using an \textbf{integrate function}, named $IF$:
\begin{equation}
    Q_1 = IF(O_{\text{flat}}, r) , \quad IF(x, y) = FC(x \oplus y \oplus ( x \otimes y) \oplus (x+y)) . \label{triliner}
\end{equation}
We add the sticker-aware utterance representation $l$ into $Q_1$ together and then apply a fully-connected layer:
\begin{align}\label{equ:q2}
    Q_2 &= FC(Q_1 \oplus l) .
\end{align}

\subsection{Fusion Network}

\begin{figure}
    \centering
    \includegraphics[scale=0.5]{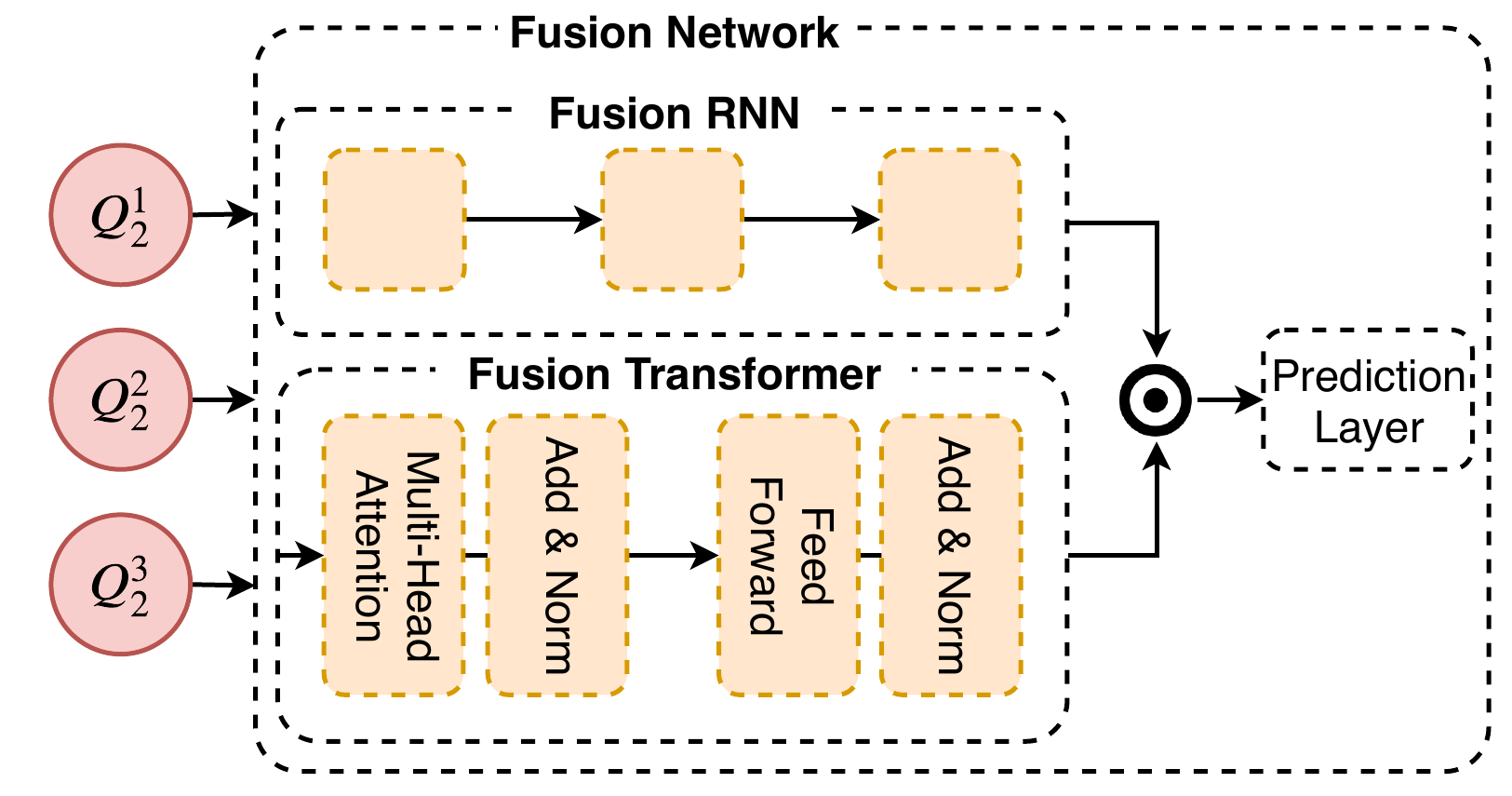}
    \caption{Framework of fusion network.}
    \label{fig:fusion}
\end{figure}

Up till now, we have obtained the interaction result between each utterance and the candidate sticker.
Here we again include the utterance index $i$ where $Q_2$ now becomes $Q_2^i$. 
Since the utterances in a multi-turn dialog context are in chronological order, we employ a \textbf{Fusion RNN} and a \textbf{Fusion Transformer} to model the short-term and long-term interaction between utterance $\{Q_2^1, \dots, Q_2^{T_u}\}$.

\subsubsection{Fusion RNN}

Fusion RNN first reads the interaction results for each utterance $\{Q_2^1, \dots, Q_2^{T_u}\}$ and then transforms into a sequence of hidden states.
In this paper, we employ the gated recurrent unit (GRU)~\cite{Chung2014EmpiricalEO} as the cell of fusion RNN, which is popular in sequential modeling~\cite{Gao2019How, Wu2017SequentialMN}:
\begin{align}
g_i &= \text{RNN}(Q_2^i, g_{i-1}) , \label{equ:fusion-rnn}
\end{align}
where $g_i$ is the hidden state of the fusion RNN.
Finally, we obtain the sequence of hidden states $\{g_1, \dots, g_{T_u}\}$.
One can replace GRU with similar algorithms such as LSTM~\cite{hochreiter1997long}.
We leave the study as future work.

\subsubsection{Fusion Transformer}
To model the long-term dependency and capture the salience utterance from the context, we employ the self-attention mechanism introduced in Equation~\ref{equ:attention}-\ref{equ:ffn}.
Concretely, given $\{Q_2^1, \dots, Q_2^{T_u}\}$, we first employ three linear projection layers with different parameters to project the input sequence into three different spaces:
\begin{equation}
\mathcal{Q}^i = FC( Q_2^i ), \quad \mathcal{K}^i = FC(Q_2^i ), \quad  \mathcal{V}^i = FC( Q_2^i).
\end{equation}
Then we feed these three matrices into the self-attention algorithm illustrated in Equation~\ref{equ:attention}-\ref{equ:ffn}.
Finally, we obtain the long-term interaction result $\{\hat{g}_1, \dots, \hat{g}_{T_u}\}$.

\subsubsection{Prediction Layer}

To combine the interaction representation generated by fusion RNN and fusion Transformer, we employ the SUMULTI function proposed by \cite{Wang2016ACM} to combine these representations, which has been proven effective in various tasks:
\begin{align}
    \overline{g}_i = \text{ReLU}(\mathcal{W}^s 
    \begin{bmatrix}
    (\hat{g}_i - g_i) \odot (\hat{g}_i - g_i) \\
    \hat{g}_i \odot g_i
   \end{bmatrix}
    + \mathbf{b}^s).
\end{align}
The new interaction sequence $\{\overline{g}_1, \dots, \overline{g}_{T_u}\}$ is then boiled down to a matching vector $\Tilde{g}_{T_u}$ by another GRU-based RNN:
\begin{align}
\Tilde{g}_i = \text{RNN}(\Tilde{g}_{i-1}, \overline{g}_i) .
\end{align}
We use the final hidden state $\Tilde{g}_{T_u}$ as the representation of the overall interaction result between the whole utterance context and the candidate sticker.
Finally, we apply a fully-connected layer to produce the matching score $\hat{y}$ of the candidate sticker:
\begin{equation}
    \hat{y} = FC(\Tilde{g}_{T_u}) ,
\end{equation}
where $\hat{y} \in (0,1)$ is the matching score of the candidate sticker.

\subsection{Learning}

Recall that we have a candidate sticker set $C=\{c_{1},...c_{T_c}\}$ which contains multiple negative samples and one ground truth sticker.
We use hinge loss as our objective function:
\begin{align}
\mathcal{L}_{r} &= \textstyle \sum^{N} \max \left( 0 , \hat{y}_{\text{negative}}- \hat{y}_{\text{positive}} +\text{margin} \right),
 \label{eq:loss-generator}
\end{align}
where $\hat{y}_{\text{negative}}$ and $\hat{y}_{\text{positive}}$ corresponds to the predicted labels of the negative sample and ground truth sticker, respectively.
The margin is the margin rescaling in hinge loss.
The gradient descent method is employed to update all the parameters in our model to minimize this loss function.
\newcommand{\cbkgrnd}{\cellcolor{blue!15}}
\section{Experimental Setup}
\label{section5}
\subsection{Research Questions}
We list four research questions that guide the experiments: 

\noindent $\bullet$ \textbf{RQ1} (See \S~\ref{subsec:Overall}): What is the overall performance of SRS compared with all baselines?

\noindent $\bullet$ \textbf{RQ2} (See \S~\ref{subsec:ablation}):  What is the effect of each module in SRS? 

\noindent $\bullet$ \textbf{RQ3} (See \S~\ref{subsec:number}): How does the performance change when the number of utterances changes?

\noindent $\bullet$ \textbf{RQ4} (See \S~\ref{subsec:attention}): 
Can co-attention mechanism successfully capture the salient part on the sticker image and the important words in dialog context? 

\noindent $\bullet$ \textbf{RQ5} (See \S~\ref{subsec:features}): What is the influence of the similarity between candidate stickers?

\noindent $\bullet$ \textbf{RQ6} (See \S~\ref{subsec:hidden}): What is the influence of the parameter settings?

\subsection{Comparison Methods}

We first conduct an ablation study to prove the effectiveness of each component in SRS as shown in Table~\ref{tab:ablations}.
Specifically, we remove each key part of our SRS to create ablation models and then evaluate the performance of these models.

\begin{table}[t]
\centering
\caption{Ablation models for comparison.}
\label{tab:ablations}
\begin{tabular}{ll}
\toprule
Acronym & Gloss \\
\midrule
SRS w/o pretrain &  \multicolumn{1}{p{5cm}}{\small SRS w/o pre-trained Inception-v3 model}\\
SRS w/o Classify &  \multicolumn{1}{p{5cm}}{\small SRS w/o emoji classification task}\\
SRS w/o DIN &  \multicolumn{1}{p{5cm}}{\small SRS w/o \textbf{D}eep \textbf{I}nteraction \text{N}etwork}\\
SRS w/o FR &  \multicolumn{1}{p{5cm}}{\small SRS w/o \textbf{F}usion \textbf{R}NN}\\

\bottomrule
\end{tabular}
\end{table}

Next, to evaluate the performance of our model, we compare it with the following baselines.
Note that, since no existing models can be directly applied to our task, we adapt VQA and multi-turn response selection models to the sticker response selection task.

\noindent (1) \textbf{Synergistic}:
\cite{guo2019image} devises a novel synergistic network on VQA task.
First, candidate answers are coarsely scored according to their relevance to the image-question pair. 
Afterward, answers with high probabilities of being correct are re-ranked by synergizing with image and question.
This model achieves the state-of-the-art performance on the Visual Dialog v1.0 dataset~\cite{das2017visual}.

\noindent (2) \textbf{PSAC}: 
\cite{li2019beyond} proposes the positional self-attention with co-attention architecture on VQA task, which does not require RNNs for video question answering. 
We replace the output probability on the vocabulary size with the probability on candidate sticker set.

\noindent (3) \textbf{SMN}: 
\cite{Wu2017SequentialMN} proposes a sequential matching network to address response selection for the multi-turn conversation problem.
SMN first matches a response with each utterance in the context.
Then vectors are accumulated in chronological order through an RNN.
The final matching score is calculated with RNN.

\noindent (4) \textbf{DAM}: 
\cite{zhou2018multi} extends the transformer model~\cite{vaswani2017attention} to the multi-turn response selection task, where representations of text segments are constructed using stacked self-attention.
Then, truly matched segment pairs are extracted across context and response. 

\noindent (5) \textbf{MRFN}: 
\cite{tao2019multi} proposes a multi-representation fusion network, where the representations can be fused into matching at an early stage, at the intermediate stage or the last stage.
This is the state-of-the-art model on the multi-turn response selection task.

For the three baselines above, we replace the candidate embedding RNN network with the image encoding CNN network Inception-v3, as used in our model.
This network is initialized using a pre-trained model\footnote{\url{https://github.com/tensorflow/models/tree/master/research/slim}} for all baselines and SRS.

\subsection{Evaluation Metrics}
Following~\cite{tao2019multi,zhou2018multi}, we employ recall at position $k$ in $n$ candidates $R_n@k$ as an evaluation metric, which measures if the positive response is ranked in the top $k$ positions of $n$ candidates.
Following~\cite{zhou2018multi}, we also employ mean average precision (MAP)~\cite{baeza2011modern} as an evaluation metric.
The statistical significance of differences observed between the performance of two runs is tested using a two-tailed paired t-test and is denoted using \dubbelop\ (or \dubbelneer) for strong significance at $\alpha=0.01$.

\subsection{Implementation Details}

We implement our experiments using TensorFlow~\cite{abadi2016tensorflow} on an NVIDIA P100 GPU. 
If the number of words in an utterance is less than 30, we pad zeros, otherwise, the first 30 words are kept.
The word embedding dimension is set to 100 and the number of hidden units is 100.
The batch size is set to 32.
9 negative samples are randomly sampled from the sticker set containing the ground truth sticker, and we finally obtain 10 candidate stickers for the model to select.
We use Adam optimizer~\cite{Kingma2015AdamAM} as our optimizing algorithm, and the learning rate is $1 \times 10^{-4}$.

\section{Experimental result}
\label{section6}

\subsection{Overall Performance}
\label{subsec:Overall}

\begin{table}[t]
\centering
\caption{RQ1: Automatic evaluation comparison. Significant differences are with respect to MRFN.}
\begin{tabular}{@{}l cc cc @{}}
\toprule
& MAP & $R_{10}@1$ & $R_{10}@2$  & $R_{10}@5$   \\
\midrule
\multicolumn{5}{@{}l}{\emph{Visual Q\&A methods}}\\
Synergistic & 0.593 \phantom{0}  & 0.438\phantom{0} & 0.569\phantom{0} & 0.798\phantom{0} \\
PSAC & 0.662\phantom{0}  & 0.533\phantom{0} & 0.641\phantom{0} & 0.836\phantom{0} \\
\midrule
\multicolumn{5}{@{}l}{\emph{Multi-turn response selection methods}}\\
SMN & 0.524\phantom{0}  & 0.357\phantom{0} & 0.488\phantom{0} & 0.737\phantom{0} \\
DAM & 0.620\phantom{0}   & 0.474\phantom{0} & 0.601\phantom{0} & 0.813\phantom{0} \\
MRFN & 0.684\phantom{0}  & 0.557\phantom{0} & 0.672\phantom{0} & 0.853\phantom{0}\\
\midrule
SRS & \textbf{0.709}  & \textbf{0.590}\dubbelop & \textbf{0.703}\dubbelop & \textbf{0.872} \\
\bottomrule
\end{tabular}
\label{tab:comp_auto_baselines}
\end{table}

\begin{figure*} 
    \centering 
    \subfigure[$MAP$ score]{ 
        \label{figs:MAP.png} 
        \includegraphics[scale=0.26]{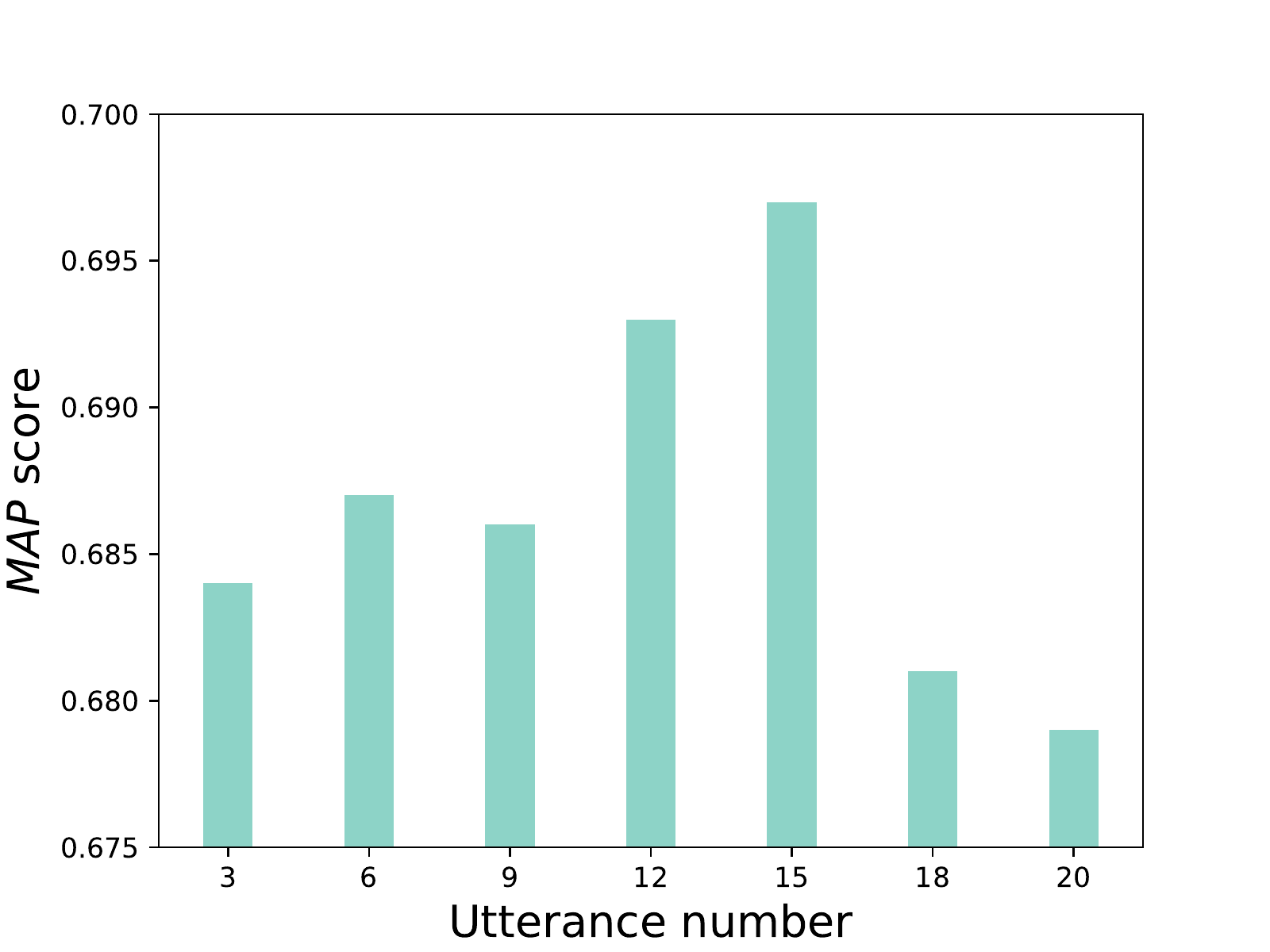}
    } 
    \subfigure[$R_{10}@1$ score]{ 
        \label{figs:r1.png} 
        \includegraphics[scale=0.26]{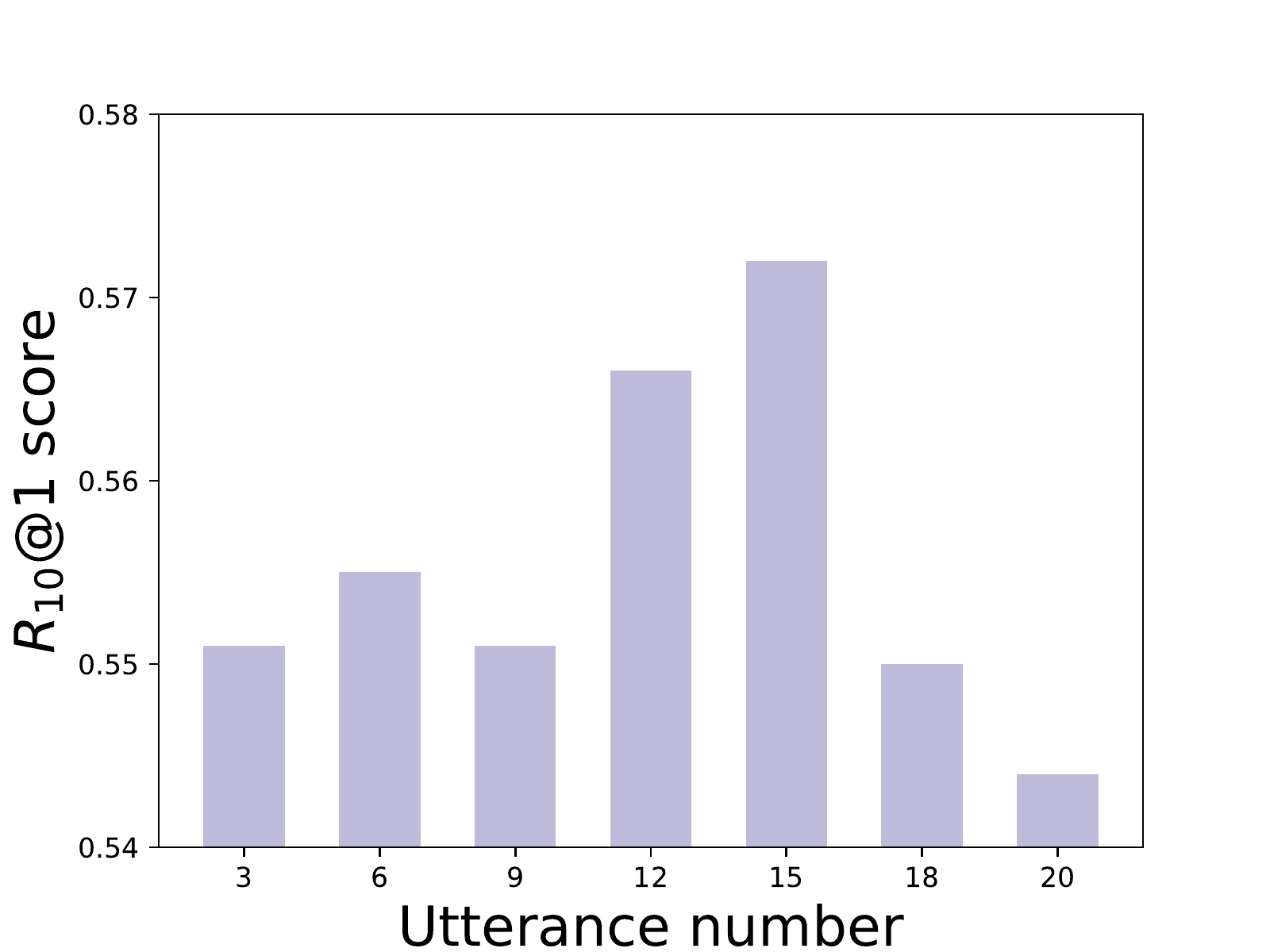}
    } 
    \subfigure[$R_{10}@2$ score]{ 
        \label{figs:r2.png} 
        \includegraphics[scale=0.26]{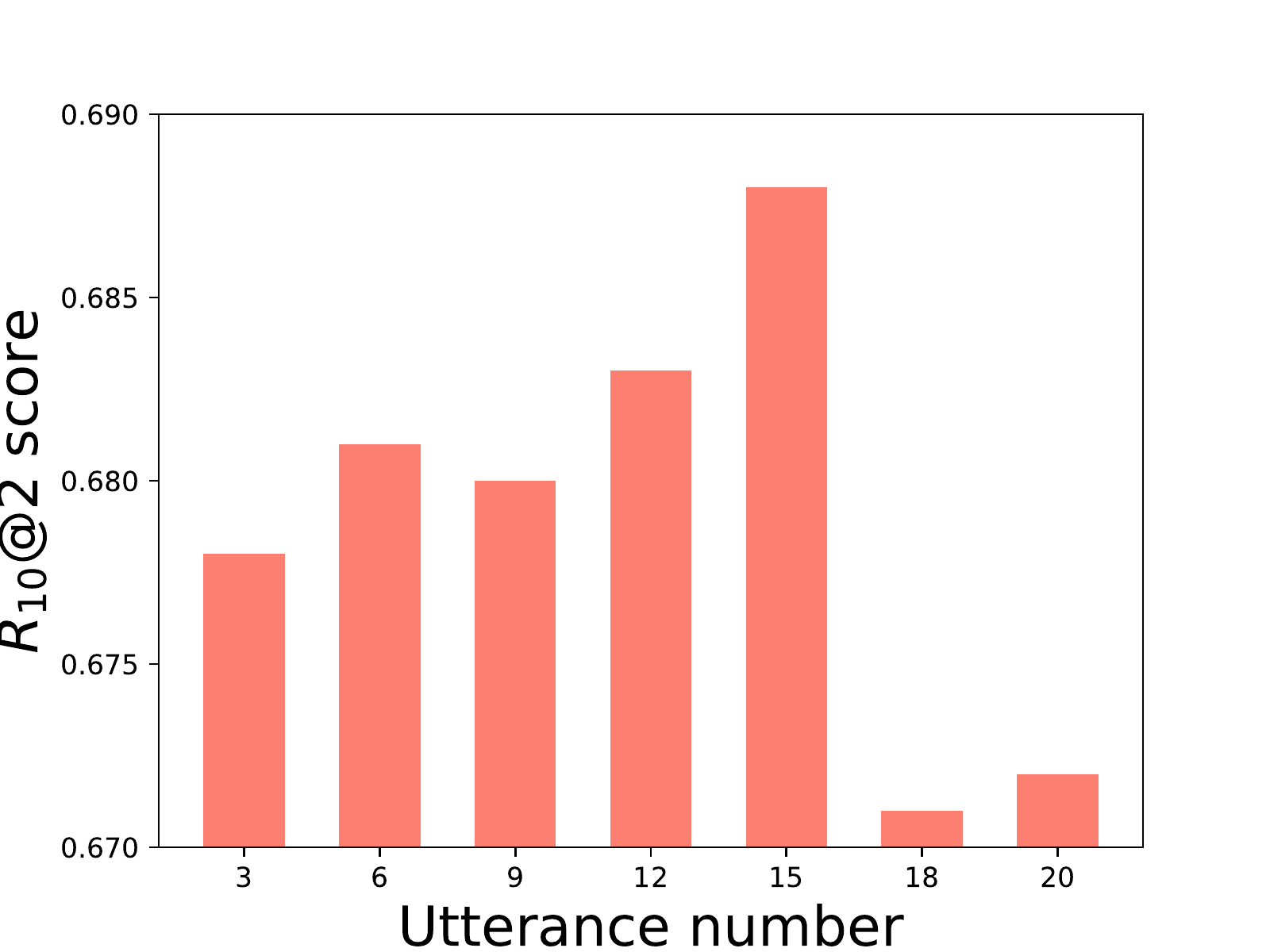}
    } 
    \subfigure[$R_{10}@5$ score]{ 
        \label{figs:r5.png} 
        \includegraphics[scale=0.26]{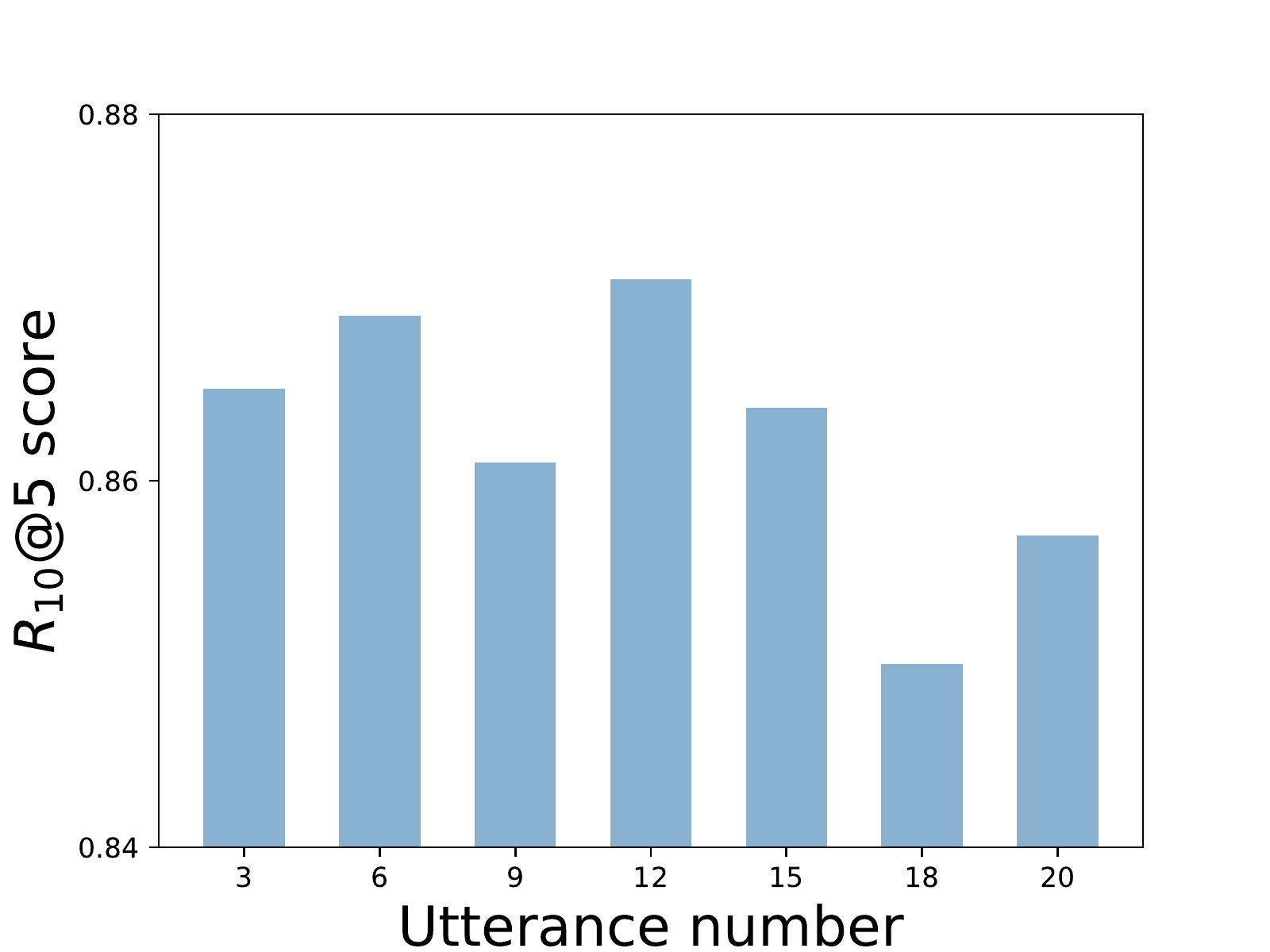}
    } 
    \caption{
         Performance of SRS on all metrics when reading different number of utterances.
    }
    \label{fig:turns}
\end{figure*}

For research question \textbf{RQ1}, we examine the performance of our model and baselines in terms of each evaluation metric, as shown in Table~\ref{tab:comp_auto_baselines}.
First, the performance of the multi-turn response selection models is generally consistent with their performances on text response selection datasets.
SMN~\cite{Wu2017SequentialMN}, an earlier work on multi-turn response selection task with a simple structure, obtains the worst performance on both sticker response and text response selection.
DAM~\cite{zhou2018multi} improves the SMN model and gets the second best performance.
MRFN~\cite{tao2019multi} is the state-of-the-art text response selection model and achieves the best performance among baselines in our task as well.
Second, VQA models perform generally worse than multi-turn response selection models, since the interaction between multi-turn utterances and sticker is important, which is not taken into account by VQA models. 
Finally, SRS achieves the best performance with 3.36\%, 5.92\% and 3.72\% improvements in MAP, $R_{10}@1$ and $R_{10}@2$ respectively, over the state-of-the-art multi-turn selection model, \ie MRFN, and with 6.80\%, 10.69\% and 8.74\% significant increases over the state-of-the-art visual dialog model, PSAC. 
This proves the superiority of our model.


\subsection{Ablation Study}
\label{subsec:ablation}

\begin{table}[t]
    \centering
    \caption{RQ2: Evaluation of different ablation models.}
    \begin{tabular}{@{}lcc cc@{}}
        \toprule
        & MAP & $R_{10}@1$ & $R_{10}@2$ & $R_{10}@5$ \\
        \midrule
        SRS w/o pretrain & 0.650 & 0.510 & 0.641 & 0.833\\
        SRS w/o Classify & 0.707 & 0.588  & 0.700 & 0.871  \\
        SRS w/o DIN & 0.680  & 0.552  & 0.669 & 0.854 \\
        SRS w/o FR & 0.677  & 0.551 & 0.663 & 0.863 \\
        SRS & \textbf{0.709}  & \textbf{0.590} & \textbf{0.703} & \textbf{0.872} \\
        \bottomrule
    \end{tabular}
    \label{tab:comp_rouge_ablation}
\end{table}

For research question \textbf{RQ2}, we conduct ablation tests on the use of the pre-trained Inception-v3 model, the sticker classification loss, the deep interaction network and the fusion RNN respectively.
The evaluation results are shown in Table~\ref{tab:comp_rouge_ablation}.
The performances of all ablation models are worse than that of SRS under all metrics, which demonstrates the necessity of each component in SRS.
We also find that the sticker classification makes the least contribution to the overall performance.
But this additional task can speed up the training process, and help our model to converge quickly.
We use 19 hours to train the SRS until convergence, and we use 30 hours for training SRS w/o Classify.
The fusion RNN brings a significant contribution, improving the MAP and $R_{10}@1$ scores by 4.43\% and 7.08\%, respectively.
Besides, the deep interaction network also plays an important part. 
Without this module, the interaction between the sticker and utterance are hindered, leading to a 6.88\% drop in $R_{10}@1$.

\subsection{Analysis of Number of Utterances} \label{subsec:number}

For research question \textbf{RQ3}, in addition to comparing with various baselines, we also evaluate our model when reading different number of utterances to study how the performance relates to number of context turns.

Figure~\ref{fig:turns} shows how the performance of the SRS changes with respect to different numbers of utterances turns.
We observe a similar trend for SRS on the first three evaluation metrics $MAP$, $R_{10}@1$ and $R_{10}@2$: they first increase until the utterance number reaches 15, and then fluctuate as the utterance number continues to increase.
There are two possible reasons for this phenomena.
The first reason might be that, when the information in the utterances is limited, the model can capture the features well, and thus when the amount of information increases, the performance gets better.
However, the capacity of the model is limited, and when the amount of information reaches its upper bound, it gets confused by this overwhelming information.
The second reason might be of the usefulness of utterance context.
Utterances that occur too early before the sticker response may be irrelevant to the sticker and bring unnecessary noise.
As for the last metric, the above observations do not preserve.
The $R_{10}@5$ scores fluctuate when the utterance number is below 15, and drop when the utterance number increases.
The reason might be that $R_{10}@5$ is not a strict metric, and it is easy to collect this right sticker in the set of half of the whole candidates.
Thus, the growth of the information given to SRS does not help it perform better but the noise it brings harms the performance.
On the other hand, though the number of utterances changes from 3 to 20, the overall performance of SRS generally remains at a high level, which proves the robustness of our model.

\begin{figure*}[h]
    \centering
    \includegraphics[scale=0.58]{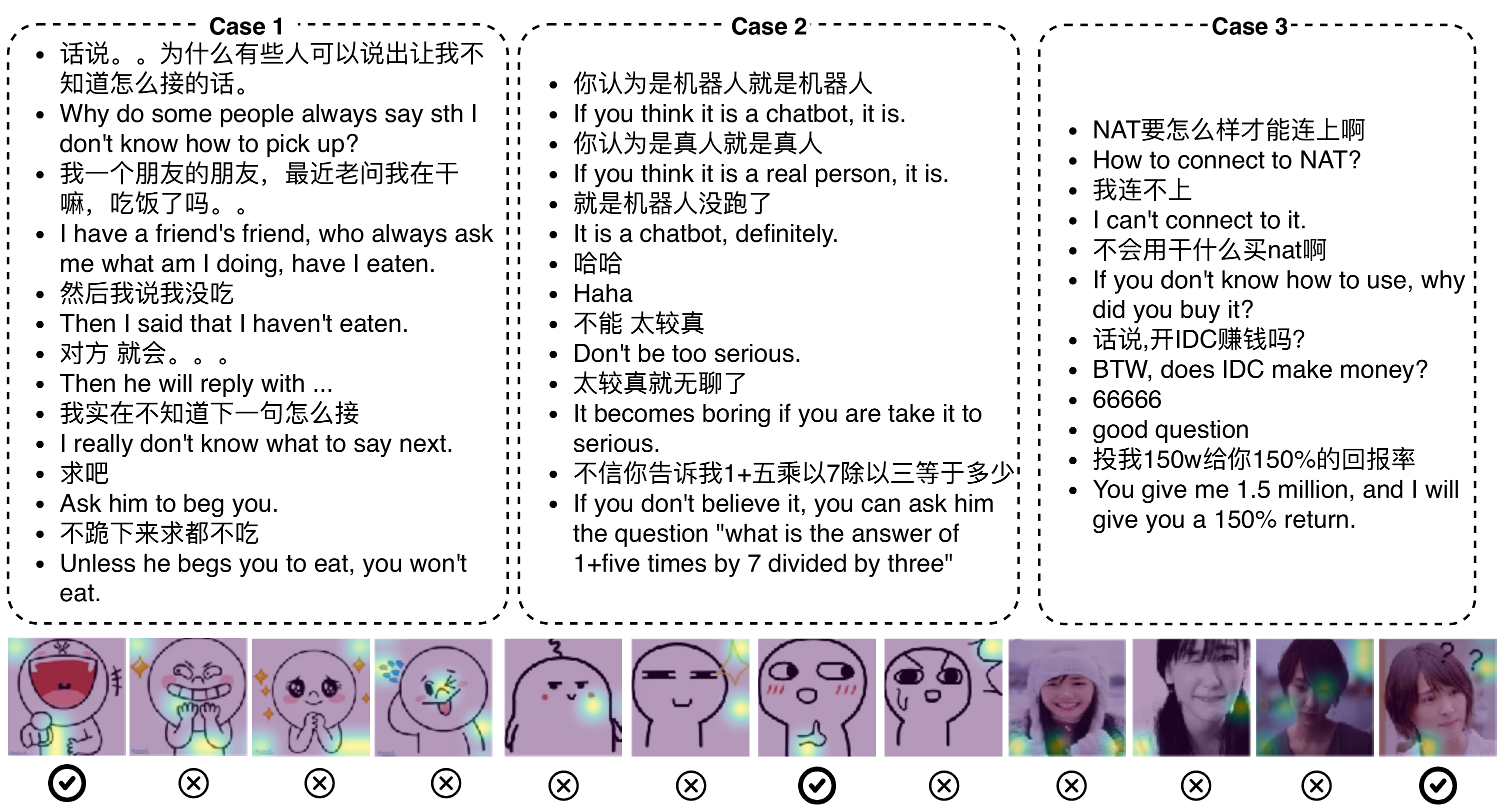}
    \caption{
        Examples of sticker selection results produced by SRS. We show the selected sticker and three random selected candidate stickers with the attention heat map. The lighter the area on image is, the higher attention weight it gets.
    }
    \label{fig:predict-case}
\end{figure*}

\begin{figure}[h]
    \centering
    \includegraphics[scale=0.55]{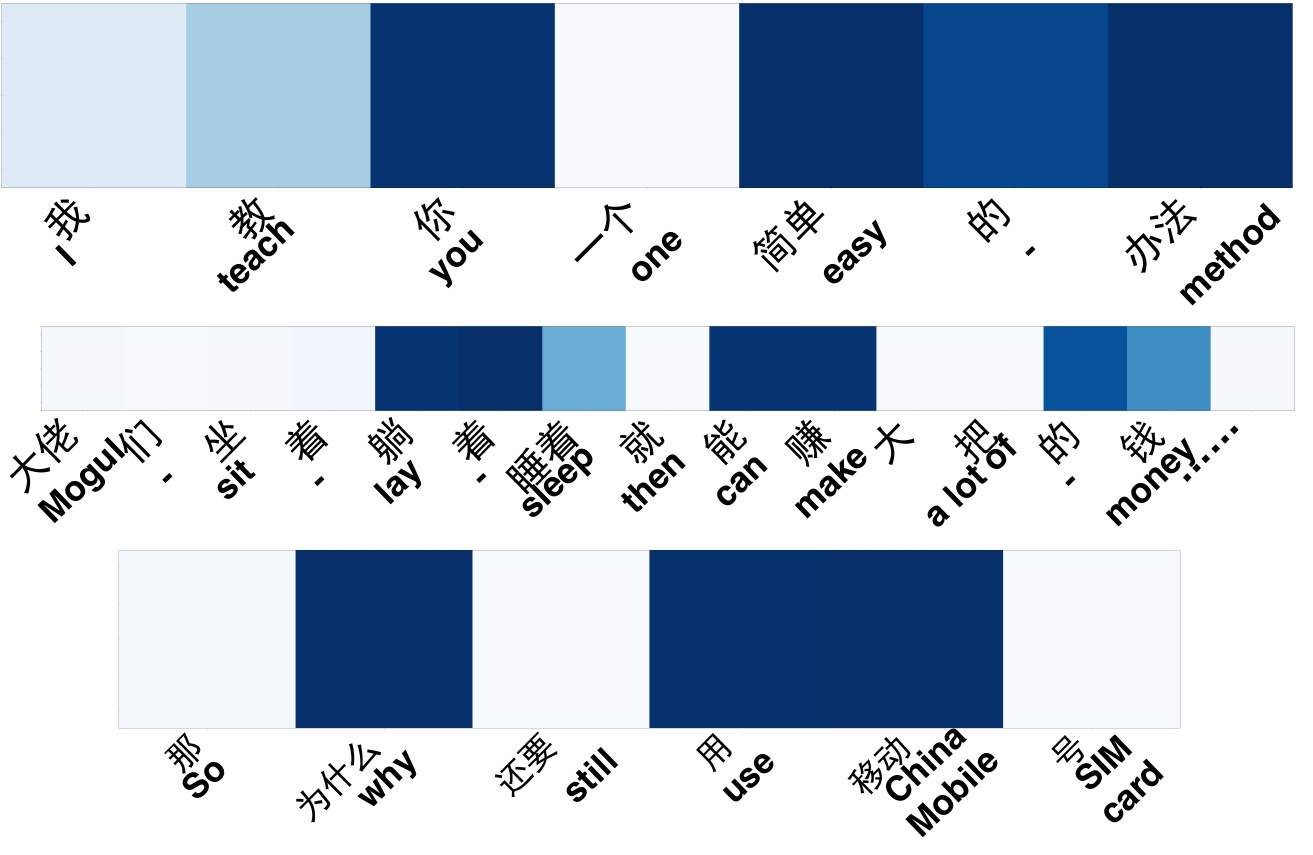}
    \caption{
        Examples of the attention weights of the dialog utterance. We translate Chinese to English word by word. The darker the area is, the higher weight the word gets.
    }
    \label{fig:text-attention-case}
\end{figure}

\subsection{Analysis of Attention Distribution in Interaction Process}
\label{subsec:attention}

Next, we turn to address \textbf{RQ4}.
We also show three cases with the dialog context in Figure~\ref{fig:predict-case}.
There are four stickers under each dialog context, one is the selected sticker by our model and other three stickers are random selected candidate stickers.
As a main component of SRS, the deep interaction network comprises a bi-directional attention mechanism between the utterance and the sticker, where each word in the utterance and each unit in the sticker representation have a similarity score in the co-attention matrix.
To visualize the sticker selection process and to demonstrate the interpretability of SRS, we visualize the sticker-wise attention $\tau^s$ (Equation~\ref{equ:ua-sticker}) on the original sticker image and show some examples in Figure~\ref{fig:predict-case}.
The lighter the area is, the higher attention it gets.

Facial expressions are an important part in sticker images.
Hence, we select several stickers with vivid facial expression in Figure~\ref{fig:predict-case}.
Take forth sticker in Case 1 for example where the character has a wink eye and a smiling mouth.
The highlights are accurately placed on the character's eye, indicating that the representation of this sticker is highly dependent on this part.
Another example is the last sticker of Case 3, there is two question marks on the top right corner of the sticker image which indicates that the girl is very suspicious of this.
In addition to facial expression, the characters gestures can also represent emotions.
Take the third sticker in Case 2 for example, the character in this sticker gives a thumbs up representing support and we can find that the attention lies on his hand, indicating that the model learns the key point of his body language.

Furthermore, we randomly select three utterances from the test dataset, and we also visualize the attention distribution over the words in a utterance, as shown in Figure~\ref{fig:text-attention-case}.
We use the weight $\tau_j^u$ for the $j$-th word (calculated in Equation~\ref{equ:sa-utterance}) as the attention weight.
We can find that the attention module always gives a higher attention weight on the salience word, such as the ``easy method'', ``make a lot of money'' and ``use Chine Mobile''.

\subsection{Influence of Similarity between Candidates}
\label{subsec:features}

\begin{figure}[h]
    \centering
    \includegraphics[scale=0.40]{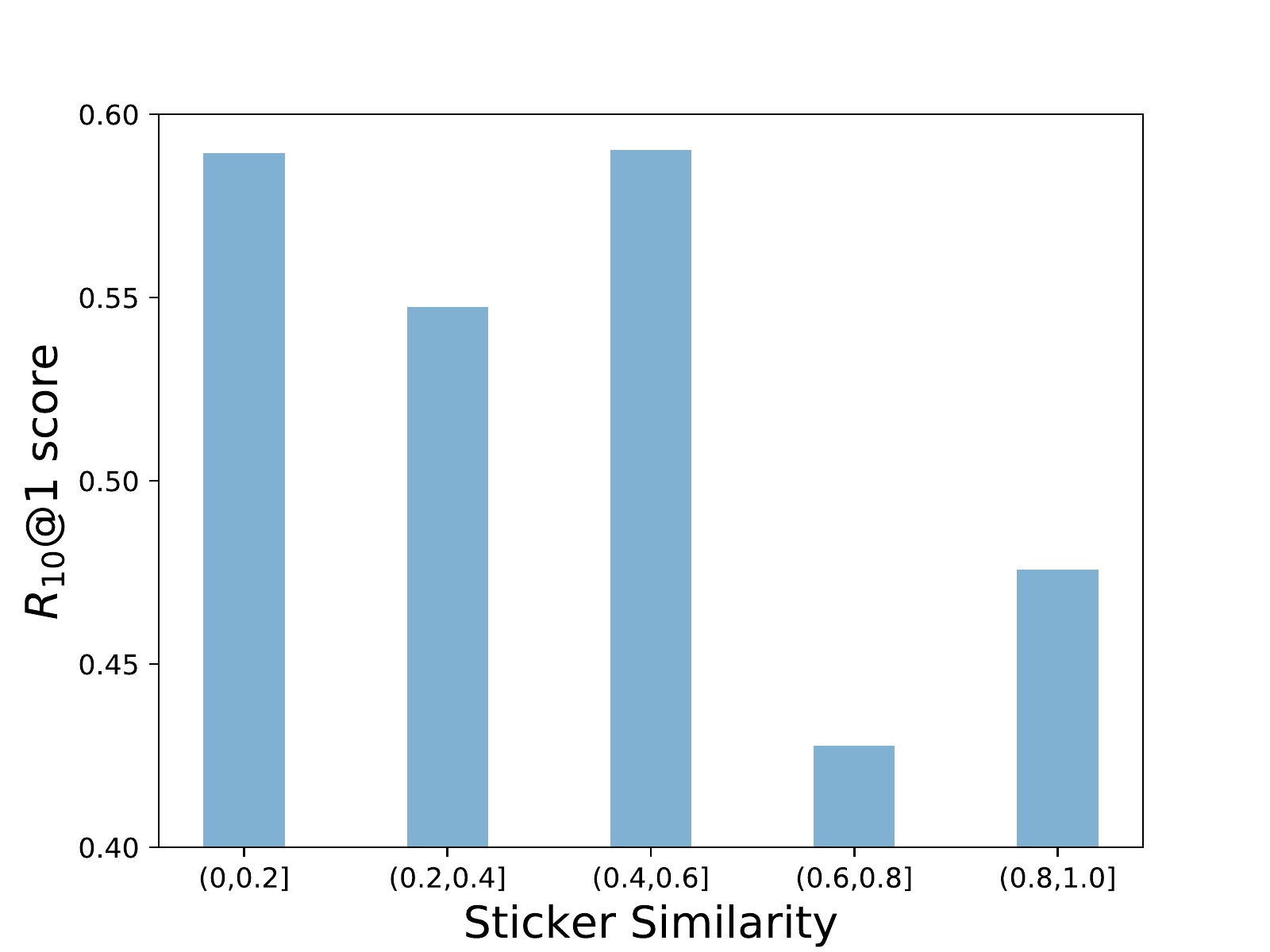}
    \caption{
        Performance of SRS on groups of different candidate similarity.
    }
    \label{fig:similarity-recall}
\end{figure}

In this section, we turn to \textbf{RQ5} to investigate the influence of the similarities between candidates.
The candidate stickers are sampled from the same set, and stickers in a set usually have a similar style.
Thus, it is natural to ask: Can our model identify the correct sticker from a set of similar candidates?
What is the influence of the similarity between candidate stickers?
Hence, we use the Structural Similarity Index (SSIM) metric~\cite{wang2004image,avanaki2008exact} to calculate the average similarity among all candidates in a test sample and then aggregate all test samples into five groups according to their average similarities.
We calculate the $R_{10}@1$ of each group of samples, as shown in Figure~\ref{fig:similarity-recall}.
The x-axis is the average similarity between candidate stickers and the y-axis is the $R_{10}@1$ score.

Not surprisingly, SRS gains the best performance when the average similarity of the candidate group is low and its performance drops as similarity increases.
However, we can also see that, though similarity varies from minimum to maximum, the overall performance can overall stay at high level. 
$R_{10}@1$ scores of all five groups are above 0.42, and the highest score reaches 0.59.
That is, our model is highly robust and can keep giving reasonable sticker responses.

\subsection{Robustness of Parameter Setting}\label{subsec:hidden}

\begin{figure}[h]
    \centering
    \includegraphics[scale=0.45]{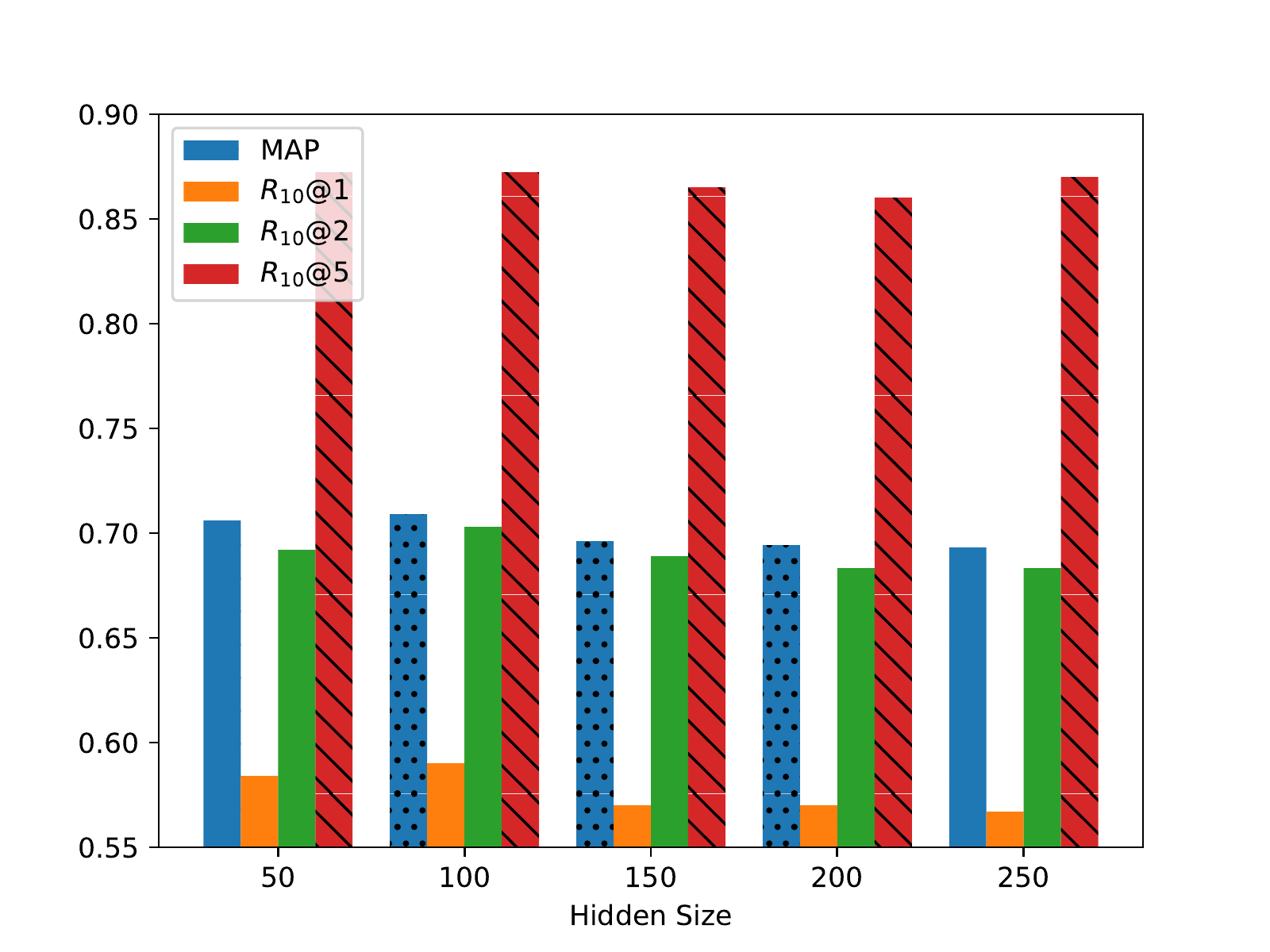}
    \caption{
        Performance of SRS with different parameter settings.
    }
    \label{fig:hidden}
\end{figure}

Finally, we turn to address \textbf{RQ6} to investigate the robustness of parameter setting.
We train our model in different parameter setting as shown in Figure~\ref{fig:hidden}.
The hidden size of the RNN, CNN and the dense layer in our model is tuned from 50 to 250, and we use the MAP and $R_n@k$ to evaluate each model.
As the hidden size grows larger from 50 to 100, the performance rises too.
The increment of hidden size improves the MAP and $R_{10}@1$ scores by 0.4\% and 1.0\%.
When the hidden size continuously goes larger from 100 to 250, the performance is declined slightly.
The increment of hidden size leads to a 2.2\% and 3.9\% drop in terms of MAP and $R_{10}@1$ respectively.
Nonetheless, we can find that each metric maintained at a stable interval, which demonstrates that our SRS is robust in terms of the parameter size.
\section{Conclusion}
\label{section7}

In this paper, we propose the task of multi-turn sticker response selection, which recommends an appropriate sticker based on multi-turn dialog context history without relying on external knowledge.
To tackle this task, we proposed the \emph{sticker response selector} (SRS).
Specifically, SRS first learns the representation of each utterance using a self-attention mechanism, and learns sticker representation by CNN.
Next, a deep interaction network is employed to fully model the dependency between the sticker and utterances.
The deep interaction network consists of a co-attention matrix that calculates the attention between each word in an utterance and each unit in a sticker representation.
Then, a bi-directional attention is used to obtain utterance-aware sticker representation and sticker-aware utterance representations.
Finally, a fusion network models the short-term and long-term relationship between interaction results, and a fully-connected layer is applied to obtain the final selection result.
Our model outperforms state-of-the-art methods in all metrics and the experimental results also demonstrate the robustness of our model on datasets with different similarity between candidate stickers.
In the near future, we aim to propose a personalized sticker response selection system.

\section*{Acknowledgments}
We would like to thank the anonymous reviewers for their constructive comments. 
We would also like to thank Anna Hennig in Inception Institute of Artificial Intelligence for her help on this paper. 
This work was supported by the National Key Research and Development Program of China (No. 2017YFC0804001), the National Science Foundation of China (NSFC No. 61876196 and NSFC No. 61672058).
Rui Yan is partially supported as a Young Fellow of Beijing Institute of Artificial Intelligence (BAAI).

\clearpage
%
\bibliographystyle{ACM-Reference-Format}
\bibliography{stickerchat}

\end{document}